\documentclass[11pt,a4paper]{article}
\usepackage[hyperref]{acl2019}
\usepackage{times}
\usepackage{latexsym}
\usepackage{graphicx}
\usepackage{url}
\usepackage{multirow}
\usepackage{amssymb}
\usepackage{algorithm,algcompatible,amsmath}
\usepackage{mathtools}
\usepackage{xcolor}
\usepackage{enumitem}
\usepackage{marvosym}
\usepackage{textcomp}
\usepackage{array}
\usepackage{graphicx}
\usepackage{url}
\usepackage{booktabs}
\newcounter{Lcount}
\newcommand{\squishenum}{
\begin{list}{\arabic{Lcount}. }
{ \usecounter{Lcount}
\setlength{\itemsep}{0pt}
\setlength{\parsep}{0pt}
\setlength{\topsep}{0pt}
\setlength{\partopsep}{0pt}
\setlength{\leftmargin}{2em}
\setlength{\labelwidth}{1.5em}
\setlength{\labelsep}{0.5em} } }
\newcommand{\squishletter}{
\begin{list}{\alph{Lcount}. }
{ \usecounter{Lcount}
\setlength{\itemsep}{0pt}
\setlength{\parsep}{0pt}
\setlength{\topsep}{0pt}
\setlength{\partopsep}{0pt}
\setlength{\leftmargin}{2em}
\setlength{\labelwidth}{1.5em}
\setlength{\labelsep}{0.5em} } }
\newcommand{\squishlist}{
\begin{list}{$\bullet$}
{ \usecounter{Lcount}
\setlength{\itemsep}{0pt}
\setlength{\parsep}{0pt}
\setlength{\topsep}{0pt}
\setlength{\partopsep}{0pt}
\setlength{\leftmargin}{2em}
\setlength{\labelwidth}{1.5em}
\setlength{\labelsep}{0.5em} } }
\newcommand{\squishend}{
\end{list} }

\aclfinalcopy 
\def\aclpaperid{1065} 

\title{Attention Guided Graph Convolutional Networks for Relation Extraction}

\author{Zhijiang Guo\thanks{$^{*}$Equally Contributed.} ~, Yan Zhang$^{*}$ \and Wei Lu \\
StatNLP Research Group\\
 Singapore University of Technology and Design \\
   \texttt{\{zhijiang\_guo,yan\_zhang\}@mymail.sutd.edu.sg, luwei@sutd.edu.sg}\\
}
\date{}

\begin{document}
\maketitle

\begin{abstract}
Dependency trees convey rich structural information that is proven useful for extracting relations among entities in text. However, how to effectively make use of relevant information while ignoring irrelevant information from the dependency trees remains a challenging research question. Existing approaches employing rule based hard-pruning strategies for selecting relevant partial dependency structures may not always yield optimal results. In this work, we propose Attention Guided Graph Convolutional Networks (AGGCNs), a novel model which directly takes full dependency trees as inputs. Our model can be understood as a soft-pruning approach that automatically learns how to selectively attend to the relevant sub-structures useful for the relation extraction task. Extensive results on various tasks including cross-sentence $n$-ary relation extraction and large-scale sentence-level relation extraction show that our model is able to better leverage the structural information of the full dependency trees, giving significantly better results than previous approaches.
\end{abstract}

\section{Introduction}
\label{sec:1}

\begin{figure*}
    \centering
    \includegraphics[scale=0.54]{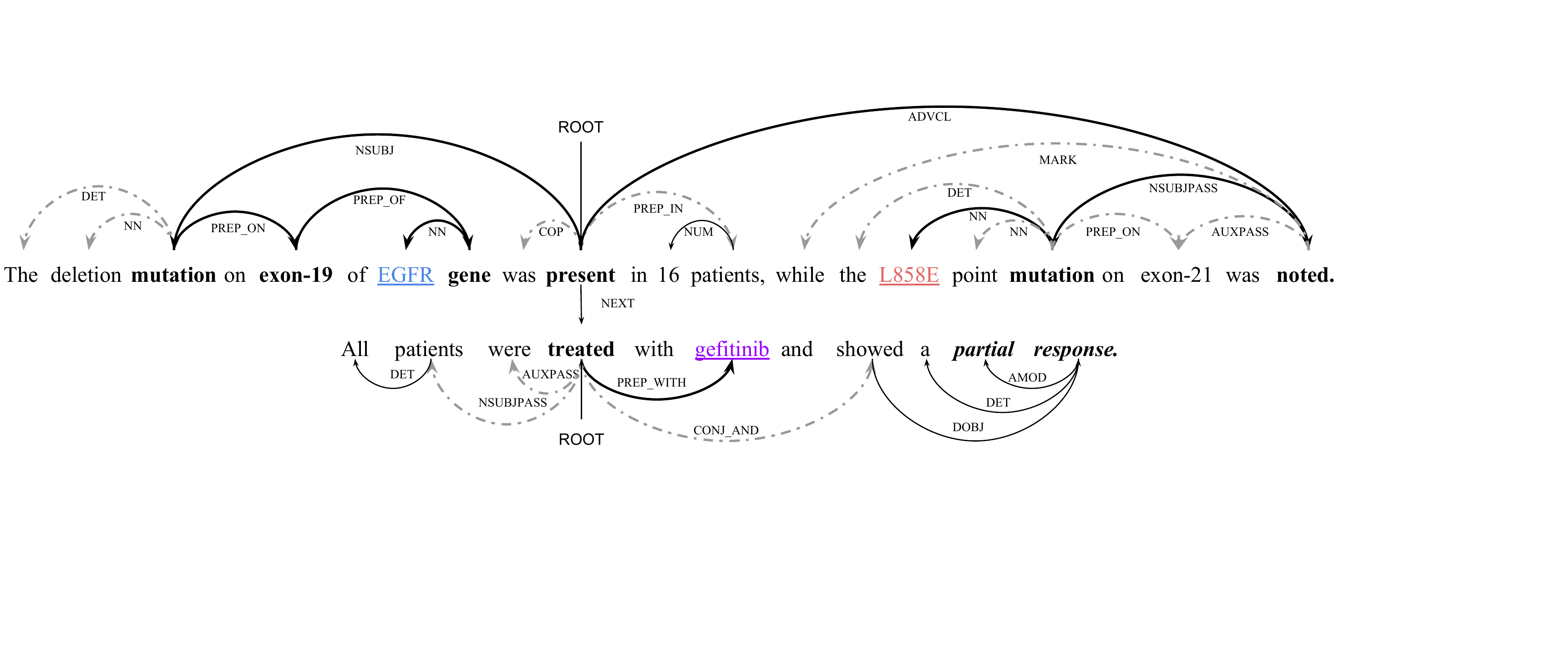}
    \caption{An example dependency tree for two sentences expressing a relation (sensitivity) among three entities. The shortest dependency path between these entities is highlighted in bold (edges and tokens). The root node of the LCA subtree of entities is \textit{present}. The dotted edges indicate tokens $K$=1 away from the subtree. Note that tokens \textit{partial response} off these paths (shortest dependency path, LCA subtree, pruned tree when $K$=1).}
    \label{fig:Figure1}
\end{figure*}

Relation extraction aims to detect relations among entities in the text. It plays a significant role in a variety of natural language processing applications including biomedical knowledge discovery \citep{Quirk2017DistantSF}, knowledge base population \citep{Zhang2017PositionawareAA} and question answering \citep{Yu2017ImprovedNR}. Figure \ref{fig:Figure1} shows an example about expressing a relation sensitivity among three entities \textit{L858E}, \textit{EGFR} and \textit{gefitinib} in two sentences.

Most existing relation extraction models can be categorized into two classes: sequence-based and dependency-based. Sequence-based models operate only on the word sequences \citep{Zeng2014RelationCV, Wang2016RelationCV}, whereas dependency-based models incorporate dependency trees into the models \citep{Bunescu2005ASP, Peng2017CrossSentenceNR}. Compared to sequence-based models, dependency-based models are able to capture non-local syntactic relations that are obscure from the surface form alone \citep{Zhang2018GraphCO}. Various pruning strategies are also proposed to distill the dependency information in order to further improve the performance. \citet{Xu2015SemanticRC, Xu2015ClassifyingRV} apply neural networks only on the shortest dependency path between the entities in the full tree. \citet{Miwa2016EndtoEndRE} reduce the full tree to the subtree below the lowest common ancestor (LCA) of the entities. \citet{Zhang2018GraphCO} apply graph convolutional networks (GCNs) \citep{Kipf2016SemiSupervisedCW} model over a pruned tree. This tree includes tokens that are up to distance $K$ away from the dependency path in the LCA subtree. 

However, rule-based pruning strategies might eliminate some important information in the full tree. Figure \ref{fig:Figure1} shows an example in cross-sentence $n$-ary relation extraction that the key tokens \textit{partial response} would be excluded if the model only takes the pruned tree into consideration.\footnote{Case study and attention visualization of this example are provided in the supplementary material.} Ideally, the model should be able to learn how to maintain a balance between including and excluding information in the full tree. In this paper, we propose the novel Attention Guided Graph Convolutional Networks (AGGCNs), which operate directly on the full tree. Intuitively, we develop a ``soft pruning'' strategy that transforms the original dependency tree into a fully connected edge-weighted graph. These weights can be viewed as the strength of relatedness between nodes, which can be learned in an end-to-end fashion by using self-attention mechanism \citep{Vaswani2017AttentionIA}.


In order to encode a large fully connected graph, we next introduce dense connections \citep{Huang2017DenselyCC} to the GCN model following \citep{dcgcnforgraph2seq19guo}. For GCNs, $L$ layers will be needed in order to capture neighborhood information that is $L$ hops away. A shallow GCN model may not be able to capture non-local interactions of large graphs. Interestingly, while deeper GCNs can capture richer neighborhood information of a graph, empirically it has been observed that the best performance is achieved with a 2-layer model \citep{Xu2018RepresentationLO}. With the help of dense connections, we are able to train the AGGCN model with a large depth, allowing rich local and non-local dependency information to be captured.

Experiments show that our model is able to achieve better performance for various tasks. For the cross-sentence relation extraction task, our model surpasses the current state-of-the-art models on multi-class ternary and binary relation extraction by 8\% and 6\% in terms of accuracy respectively. For the large-scale sentence-level extraction task (TACRED dataset), our model is also consistently better than others, showing the effectiveness of the model on a large training set. Our code is available at \url{https://github.com/Cartus/AGGCN_TACRED}\footnote{Implementation is based on Pytorch \citep{Paszke2017AutomaticDI}.}

Our contributions are summarized as follows:

\squishlist
\item We propose the novel AGGCNs that learn a ``soft pruning'' strategy in an end-to-end fashion, which learns how to select and discard information. Combining with dense connections, our AGGCN model is able to learn a better graph representation.
\item Our model achieves new state-of-the-art results without additional computational overhead when compared with previous GCNs.\footnote{The size of the adjacency matrix representing the fully connected graph is the same as the one of the original tree.} Unlike tree-structured models (e.g., Tree-LSTM \citep{Tai2015ImprovedSR}), it can be efficiently applied over dependency trees in parallel.
\squishend

\section{Attention Guided GCNs}
\label{sec:2}

In this section, we will present the basic components used for constructing our AGGCN model. 

\begin{figure*}
    \centering
    \includegraphics[scale=0.6]{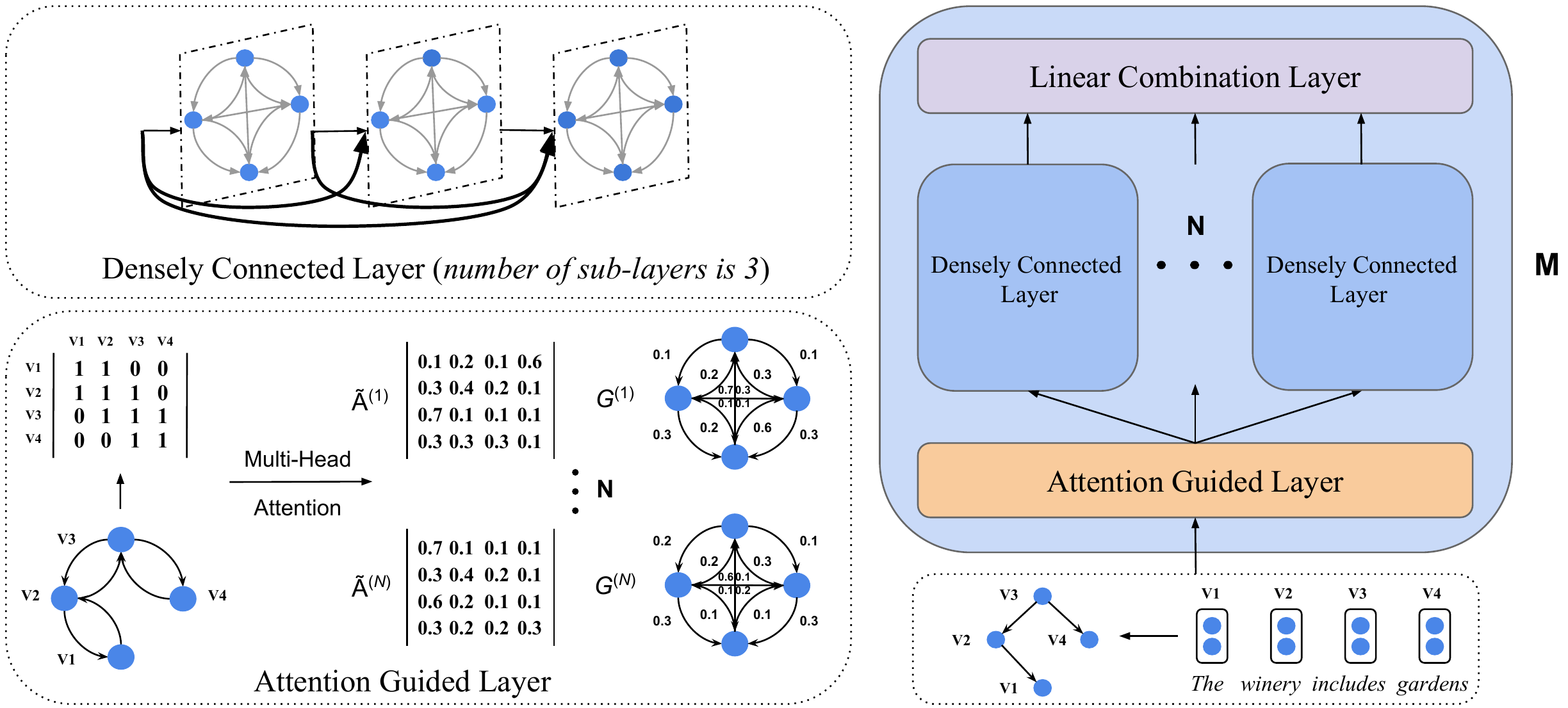}
    \caption{The AGGCN model is shown with an example sentence and its dependency tree. It is composed of $M$ identical blocks and each block has three types of layers as shown on the right. Every block takes node embeddings and adjacency matrix that represents the graph as inputs. Then $N$ attention guided adjacency matrices are constructed by using multi-head attention as shown at bottom left. The original dependency tree is transformed into $N$ different fully connected edge-weighted graphs (self-loops are omitted for simplification). Numbers near the edges represent the weights in the matrix. Resulting matrices are fed into $N$ separate densely connected layers, generating new representations. Top left shows an example of the densely connected layer, where the number ($L$) of sub-layers is 3 ($L$ is a hyper-parameter). Each sub-layer concatenates all preceding outputs as the input. Eventually, a linear combination is applied to combine outputs of $N$ densely connected layers into hidden representations.}
    \label{fig:Figure2}
\end{figure*}

\subsection{GCNs}
\label{ssec:2.1}

GCNs are neural networks that operate directly on graph structures \citep{Kipf2016SemiSupervisedCW}.  Here we mathematically illustrate how multi-layer GCNs work on a graph. Given a graph with $n$ nodes, we can represent the graph with an $n \times n$ adjacency matrix $\mathbf{A}$. \citet{Marcheggiani2017EncodingSW} extend GCNs for encoding dependency trees by incorporating directionality of edges into the model. They add a self-loop for each node in the tree. Opposite direction of a dependency arc is also included, which means $\mathbf{A}_{ij}=1$ and $\mathbf{A}_{ji}=1$ if there is an edge going from node $i$ to node $j$, otherwise $\mathbf{A}_{ij}=0$ and $\mathbf{A}_{ji}=0$. The convolution computation for node $i$ at the $l$-th layer, which takes the input feature representation $\mathbf{h}^{(l-1)}$ as input and outputs the induced representation $\mathbf{h}_i^{(l)}$, can be defined as:
\begin{equation}
\mathbf{h}_{i}^{(l)} = \rho \Big(\sum_{j=1}^{n} \mathbf{A}_{ij} \mathbf{W}^{(l)} \mathbf{h}_{j}^{(l-1)} + \mathbf{b}^{(l)} \Big)
\end{equation}
where $\mathbf{W}^{(l)}$ is the weight matrix, $\mathbf{b}^{(l)}$ is the bias vector,  and $\rho$ is an activation function (e.g., RELU). $\mathbf{h}^{(0)}_i$ is the initial input $\mathbf{x}_i$, where $\mathbf{x}_i \in \mathbb{R}^{d}$ and $d$ is the input feature dimension.

\subsection{Attention Guided Layer}
\label{ssec:2.2}
The AGGCN model is composed of $M$ identical blocks as shown in Figure \ref{fig:Figure2}. Each block consists of three types of layers: attention guided layer, densely connected layer and linear combination layer. We first introduce the attention guided layer of the AGGCN model.


As we discuss in Section \ref{sec:1}, most existing pruning strategies are predefined. They prune the full tree into a subtree, based on which the adjacency matrix is constructed. In fact, such strategies can also be viewed as a form of hard attention \citep{Xu2015ShowAA}, where edges that connect nodes not on the resulting subtree will be directly assigned zero weights (not attended). Such strategies might eliminate relevant information from the original dependency tree. Instead of using rule-based pruning, we develop a ``soft pruning'' strategy in the attention guided layer, which assigns weights to all edges. These weights can be learned by the model in an end-to-end fashion.

In the attention guided layer, we transform the original dependency tree into a fully connected edge-weighted graph by 
constructing an attention guided adjacency matrix $\mathbf{\tilde{A}}$. Each $\mathbf{\tilde{A}}$ corresponds to a certain fully connected graph and each entry $\mathbf{\tilde{A}}_{ij}$ is the weight of the edge going from node $i$ to node $j$. As shown in Figure \ref{fig:Figure2}, $\mathbf{\tilde{A}^{(1)}}$ represents a fully connected graph $G^{(1)}$. $\mathbf{\tilde{A}}$ can be constructed by using self-attention mechanism \citep{Cheng2016LongSM}, which is an attention mechanism \citep{Bahdanau2015NeuralMT} that captures the interactions between two arbitrary positions of a single sequence. Once we get $\mathbf{\tilde{A}}$, we can use it as the input for the computation of the later graph convolutional layer. Note that the size of $\mathbf{\tilde{A}}$ is the same as the original adjacency matrix $\mathbf{A}$ ($n \times n$). Therefore, no additional computational overhead is involved. The key idea behind the attention guided layer is to use attention for inducing relations between nodes, especially for those connected by indirect, multi-hop paths. These soft relations can be captured by differentiable functions in the model. 

Here we compute $\mathbf{\tilde{A}}$ by using multi-head attention \citep{Vaswani2017AttentionIA}, which allows the model to jointly attend to information from different representation subspaces. The calculation involves a query and a set of key-value pairs. The output is computed as a weighted sum of the values, where the weight is computed by a function of the query with the corresponding key.

\begin{equation}
\mathbf{\tilde{A}^{(t)}} = softmax(\frac{Q\mathbf{W}_{i}^{Q} \times (K\mathbf{W}_{i}^{K})^{T}}{\sqrt{d}})
\end{equation}
where $Q$ and $K$ are both equal to the collective representation $\mathbf{h}^{(l-1)}$ at layer $l-1$ of the AGGCN model. The projections are parameter matrices $\mathbf{W}_{i}^{Q} \in \mathbb{R}^{d \times d}$ and $\mathbf{W}_{i}^{K} \in \mathbb{R}^{d \times d}$. $\mathbf{\tilde{A}^{(t)}}$ is the $t$-th attention guided adjacency matrix corresponding to the $t$-th head. Up to $N$ matrices are constructed, where $N$ is a hyper-parameter.

Figure \ref{fig:Figure2} shows an example that the original adjacency matrix is transformed into multiple attention guided adjacency matrices. Accordingly, the input dependency tree is converted into multiple fully connected edge-weighted graphs. In practice, we treat the original adjacency matrix as an initialization so that the dependency information can be captured in the node representations for later attention calculation. The attention guided layer is included starting from the second block.

\subsection{Densely Connected Layer}
\label{ssec:2.3}

Unlike previous pruning strategies, which lead to a resulting structure that is smaller than the original structure, our attention guided layer outputs a larger fully connected graph. Following \citep{dcgcnforgraph2seq19guo}, we introduce dense connections \citep{Huang2017DenselyCC} into the AGGCN model in order to capture more structural information on large graphs. With the help of dense connections, we are able to train a deeper model, allowing rich local and non-local information to be captured for learning a better graph representation.

Dense connectivity is shown in Figure \ref{fig:Figure2}. Direct connections are introduced from any layer to all its preceding layers. Mathematically, we first define $\mathbf{g}_{j}^{(l)}$ as the concatenation of the initial node representation and the node representations produced in layers $1$, $\cdots$, $l-1$:
\begin{equation}
\mathbf{g}_{j}^{(l)} = [\mathbf{x}_{j};\mathbf{h}_{j}^{(1)}; ... ;\mathbf{h}_{j}^{(l-1)}]. 
\end{equation}

In practice, each densely connected layer has $L$ sub-layers. The dimensions of these sub-layers $d_{hidden}$ are decided by $L$ and the input feature dimension $d$. In AGGCNs, we use $d_{hidden} = d/L$. For example, if the densely connected layer has 3 sub-layers and the input dimension is 300, the hidden dimension of each sub-layer will be $d_{hidden} = d/L = 300/3=100$. Then we concatenate the output of each sub-layer to form the new representation. Therefore, the output dimension is 300 (3 $\times$ 100). Different from the GCN model whose hidden dimension is larger than or equal to the input dimension, the AGGCN model shrinks the hidden dimension as the number of layers increases in order to improves the parameter efficiency similar to DenseNets \citep{Huang2017DenselyCC}.   

Since we have $N$ different attention guided adjacency matrices, $N$ separate densely connected layers are required. Accordingly, we modify the computation of each layer as follows (for the $t$-th matrix $\mathbf{\tilde{A}^{(t)}}$):
\begin{equation}
\mathbf{h}_{t_{i}}^{(l)} = \rho \Big(\sum_{j=1}^{n} \mathbf{\tilde{A}}_{ij}^{(t)} \mathbf{W}_{t}^{(l)} \mathbf{g}_{j}^{(l)} + \mathbf{b}_{t}^{(l)} \Big)
\end{equation}
where $t= 1, ..., N$ and $t$ selects the weight matrix and bias term associated with the attention guided adjacency matrix $\mathbf{\tilde{A}^{(t)}}$. The column dimension of the weight matrix increases by $d_{hidden}$ per sub-layer, i.e., $\mathbf{W}_{t}^{(l)} \in \mathbb{R}^{d_{hidden} \times d^{(l)}}$, where $ d^{(l)} = d+d_{hidden} \times (l-1)$.

\subsection{Linear Combination Layer}
\label{ssec:2.4}

The AGGCN model includes a linear combination layer to integrate representations from $N$ different densely connected layers. Formally, the output of the linear combination layer is defined as:
\begin{equation}
\mathbf{h}_{comb} = \mathbf{W}_{comb}\mathbf{h}_{out} + \mathbf{b}_{comb}
\end{equation}
where $\mathbf{h}_{out}$ is the output by concatenating outputs from $N$ separate densely connected layers, i.e., $\mathbf{h}_{out} = [\mathbf{h}^{(1)}; ... ;\mathbf{h}^{(N)}] \in \mathbb{R}^{d \times N}$. $\mathbf{W}_{comb} \in \mathbb{R}^{(d \times N) \times d}$ is a weight matrix and $\mathbf{b}_{comb}$ is a bias vector for the linear transformation.

\subsection{AGGCNs for Relation Extraction}
\label{ssec:2.5}

After applying the AGGCN model over the dependency tree, we obtain hidden representations of all tokens. Given these representations, the goal of relation extraction is to predict a relation among entities. Following \citep{Zhang2018GraphCO}, we concatenate the sentence representation and entity representations to get the final representation for classification. First we need to obtain the sentence representation $h_{sent}$. It can be computed as:
\begin{equation}
h_{sent} = f(\mathbf{h_{mask}}) = f(\text{AGGCN}(\mathbf{x}))
\end{equation}
where $\mathbf{h_{mask}}$ represents the masked collective hidden representations. 
Masked here means we only select representations
of tokens that are not entity tokens in the sentence. $f: \mathbb{R}^{d \times n} \rightarrow \mathbb{R}^{d \times 1}$ is a max pooling function that maps from $n$ output vectors to 1 sentence vector. Similarly, we can obtain the entity representations. For the $i$-th entity, its representation $h_{e_{i}}$ can be computed as:
\begin{equation}
h_{e_{i}} = f(\mathbf{h_{e_{i}}})
\end{equation}
where $\mathbf{h_{e_{i}}}$ indicates the hidden representation corresponding to the $i$-th entity.\footnote{The number of entities is fixed in $n$-ary relation extraction task. It is 3 for the first dataset and 2 for the second.} Entity representations will be concatenated with sentence representation to form a new representation. Following \citep{Zhang2018GraphCO}, we apply a feed-forward neural network (FFNN) over the concatenated representations inspired by relational reasoning works \citep{Santoro2017ASN, Lee2017EndtoendNC}:
\begin{equation}
h_{final} = \text{FFNN}([h_{sent};h_{e_{1}};...h_{e_{i}}])
\end{equation}
where $h_{final}$ will be taken as inputs to a logistic regression classifier to make a prediction.

\section{Experiments}

\subsection{Data}
We evaluate the performance of our model on two tasks, namely,  cross-sentence $n$-ary relation extraction and sentence-level relation extraction. 

For the cross-sentence $n$-ary relation extraction task, we use the dataset introduced in ~\citep{Peng2017CrossSentenceNR}, which contains 6,987 ternary relation instances and 6,087 binary relation instances extracted from PubMed.\footnote{The dataset is available at \url{https://github.com/freesunshine0316/nary-grn}\label{n-ary_data}} Most instances contain multiple sentences and each instance is assigned with one of the five labels, including: ``resistance or nonresponse'', ``sensitivity'', ``response'', ``resistance''
and ``None''. We consider two specific tasks for evaluation, i,e., binary-class $n$-ary relation extraction and multi-class $n$-ary relation extraction.  For binary-class $n$-ary relation extraction, we follow~\citep{Peng2017CrossSentenceNR} to binarize multi-class labels by grouping the four relation classes as ``Yes'' and treating ``None'' as ``No''.  


For the sentence-level relation extraction task,  we follow the experimental settings in ~\citep{Zhang2018GraphCO} to evaluate our model on the TACRED dataset ~\citep{Zhang2017PositionawareAA} and Semeval-10 Task 8 ~\citep{Hendrickx2010SemEval2010T8}. With over 106K instances, the TACRED dataset introduces 41 relation types and a special ``no relation'' type to describe the relations between the mention pairs in instances. Subject mentions are categorized into person and organization, while object mentions are categorized into 16 fine-grained types, including date, location, etc. Semeval-10 Task 8 is a public dataset, which contains 10,717 instances with 9 relations and a special ``other'' class. 

\begin{table*}[!th]
\centering
\setlength{\tabcolsep}{2.5pt}
\scalebox{0.9}{
\begin{tabular}{lcccccccc} 
\toprule
    \multirow{3}{*}{\textbf{Model}} & \multicolumn{4}{c}{\textbf{Binary-class}} & \multicolumn{2}{c}{\textbf{Multi-class}} \\
\cmidrule(l{5pt}r{5pt}){2-5} \cmidrule(l{5pt}r{5pt}){6-7}
    & \multicolumn{2}{c}{T} & \multicolumn{2}{c}{B}  & T & B\\
& \texttt{Single} & \texttt{Cross} & \texttt{Single} & \texttt{Cross} & \texttt{Cross} & \texttt{Cross}  \\
\midrule
Feature-Based~\citep{Quirk2017DistantSF}        & 74.7 & 77.7 & 73.9 & 75.2  &- & -\\
SPTree~\citep{Miwa2016EndtoEndRE}            &  -   & -    &75.9  & 75.9  &- & -\\
Graph LSTM-EMBED~\citep{Peng2017CrossSentenceNR}& 76.5 & 80.6 &74.3  & 76.5 &- & -\\
Graph LSTM-FULL~\citep{Peng2017CrossSentenceNR} & 77.9 &80.7  &75.6  & 76.7 &- & -\\
{\color{white}00000000000000000} + multi-task                      & -    &  82.0  & -  & 78.5 &- & -\\
Bidir DAG LSTM~\citep{Song2018NaryRE}           & 75.6 & 77.3 &76.9  & 76.4    & 51.7 &	50.7\\
GS GLSTM~\citep{Song2018NaryRE}                 & 80.3 & 83.2 &83.5  & 83.6    & 71.7 &	71.7 \\
\midrule
GCN (Full Tree)~\citep{Zhang2018GraphCO}                                   &84.3 &84.8 & 84.2 & 83.6 & 77.5  &	74.3\\
GCN ($K$=0)~\citep{Zhang2018GraphCO}                                             &85.8    &85.8 & 82.8 & 82.7 & 75.6  &	72.3\\
GCN ($K$=1)~\citep{Zhang2018GraphCO}                                        &85.4    &85.7   & 83.5 & 83.4 & 78.1  &	73.6\\
GCN ($K$=2)~\citep{Zhang2018GraphCO}                                        &84.7    &85.0   & 83.8 & 83.7 & 77.9  &	73.1\\
\midrule
AGGCN (ours)                                     & \textbf{87.1} & \textbf{87.0} & \textbf{85.2} & \textbf{85.6}  & \textbf{79.7} &  \textbf{77.4}\\
\bottomrule
\end{tabular}}
\caption{Average test accuracies in five-fold validation for binary-class $n$-ary relation extraction and multi-class $n$-ary relation extraction.  ``T'' and  ``B'' denote ternary drug-gene-mutation interactions and binary drug-mutation interactions, respectively.  \texttt{Single} means that we report the accuracy on instances within single sentences, while \texttt{Cross} means the accuracy on all instances. $K$ in the GCN models means that the preprocessed pruned trees include tokens up to distance $K$ away from the dependency path in the  LCA subtree.}
\vspace{-3mm}
\label{tab:tern_bin}
\end{table*}

\subsection{Setup}
We tune the hyper-parameters according to results on the development sets. For the cross-sentence $n$-ary relation extraction task, we use the same data split  used in \citep{Song2018NaryRE}\textsuperscript{\ref{n-ary_data}}, while for the sentence-level relation extraction task, we use the same development set from~\citep{Zhang2018GraphCO}.

We choose the number of heads $N$ for attention guided layer from $\{1,2,3,4\}$, the block number $M$ from $\{1,2,3\}$, the number of sub-layers $L$ in each densely connected layer from $\{2,3,4\}$. Through preliminary experiments on the development sets, we find that the combinations ($N$=2, $M$=2, $L_{1}$=2, $L_{2}$=4, $d_{hidden}$=340)\footnote{Similar to ~\citep{dcgcnforgraph2seq19guo}, each layer has two different sub-layers.} and ($N$=3, $M$=2, $L_{1}$=2, $L_{2}$=4, $d_{hidden}$=300) give the best results on cross-sentence $n$-ary relation extraction and sentence-level relation extraction, respectively. GloVe~\citep{Pennington2014GloveGV} vectors are used as the initialization for word embeddings.

Models are evaluated using the same metrics as previous work~\citep{Song2018NaryRE, Zhang2018GraphCO}. We report the test accuracy averaged over five cross validation folds~\citep{Song2018NaryRE} for the cross-sentence $n$-ary relation extraction task. For the sentence-level relation extraction task, we report the micro-averaged F1 scores for the TACRED dataset and the macro-averaged F1 scores for the SemEval dataset~\citep{Zhang2018GraphCO}. For TACRED dataset, we report the mean test F1 score by using 5 models from independent runs.

\subsection{Results on Cross-Sentence $n$-ary Relation Extraction}
For cross-sentence $n$-ary relation extraction task, we consider three kinds of models as baselines: 1) a feature-based classifier~\citep{Quirk2017DistantSF} based on shortest dependency paths between all entity pairs, 2) Graph-structured LSTM methods, including Graph LSTM~\citep{Peng2017CrossSentenceNR}, bidirectional DAG LSTM (Bidir DAG LSTM) ~\citep{Song2018NaryRE} and  Graph State LSTM (GS GLSTM) ~\citep{Song2018NaryRE}. These methods  extend LSTM to encode graphs constructed from input sentences with dependency edges,  3) Graph  convolutional  networks (GCN) with pruned trees,  which have shown efficacy on the  relation extraction task~\citep{Zhang2018GraphCO}\footnote{The results are produced by the open implementation of~\citet{Zhang2018GraphCO}.}.  Additionally, we follow~\citep{Song2018NaryRE} to consider the tree-structured LSTM method (SPTree)~\citep{Miwa2016EndtoEndRE} on drug-mutation binary relation extraction. Main results are shown in Table~\ref{tab:tern_bin}.

We first focus on the binary-class $n$-ary relation extraction task.  For ternary relation extraction (first two columns in Table~\ref{tab:tern_bin} ), our AGGCN model achieves accuracies of 87.1 and 87.0  on instances within single sentence  ($\texttt{Single}$) and on all instances ($\texttt{Cross}$), respectively,  which outperform all the baselines. More specifically, our AGGCN model surpasses the state-of-the-art     Graph-structured LSTM model (GS GLSTM) by 6.8 and 3.8 points for the $\texttt{Single}$ and $\texttt{Cross}$ settings, respectively. Compared to GCN models , our model   obtains 1.3 and 1.2 points higher than the best performing model with pruned tree ($K$=1). For binary relation extraction (third and fourth columns in Table~\ref{tab:tern_bin}), AGGCN consistently outperforms GS GLSTM and GCN as well. 

These results suggest that, compared to previous full tree based methods, e.g., GS GLSTM,  AGGCN is able to extract more information from the underlying graph structure to learn a more expressive representation through graph convolutions. AGGCN also performs better than GCNs, although its performance can be boosted via pruned trees. We believe this is because of the combination of densely connected layer and attention guided layer. The dense connections could facilitate information propagation in large graphs, enabling AGGCN to efficiently learn from long-distance dependencies without pruning techniques. Meanwhile, the attention guided layer can further distill relevant information and filter out noises from the representation learned by the densely connected layer. 

We next show the results on the multi-class classification task (last two columns in Table~\ref{tab:tern_bin}). We follow~\citep{Song2018NaryRE} to evaluate our model on all instances for both ternary and binary relations. This fine-grained classification task is much harder than coarse-grained classification task. As a result,  the performance of all models degrades a lot. However, our AGGCN model still obtains 8.0 and 5.7  points  higher than the GS GLSTM model for ternary and binary relations, respectively. We also notice that our AGGCN achieves a better test accuracy than all GCN models, which further demonstrates its ability to learn better representations from full trees. 

\subsection{Results on Sentence-level Relation Extraction}

We now report the results on the TACRED dataset for the sentence-level relation extraction task  in Table~\ref{tab:tacred}.  We compare our model against two kinds of  models: 1) dependency-based models, 2) sequence-based models. Dependency-based models include the logistic regression classifier (LR) ~\citep{Zhang2017PositionawareAA}, Shortest Path LSTM (SDP-LSTM)~\citep{Xu2015ClassifyingRV}, Tree-structured neural model (Tree-LSTM)~\citep{Tai2015ImprovedSR}, GCN and Contextualized GCN (C-GCN) ~\citep{Zhang2018GraphCO}.   Both GCN and C-GCN models use the pruned trees. For sequence-based model, we consider the state-of-the-art  Position Aware LSTM (PA-LSTM)~\citep{Zhang2017PositionawareAA}.



\begin{table}[!t]
	\centering
\setlength{\tabcolsep}{3pt}
\begin{tabular}{lccc}
\toprule
\bf Model & P & R & F1 \\
\midrule
LR~\citep{Zhang2017PositionawareAA}          & \textbf{73.5} &49.9 &	59.4 \\
SDP-LSTM~\citep{Xu2015ClassifyingRV}{*}    & 66.3 &52.7 &	58.7 \\
Tree-LSTM~\citep{Tai2015ImprovedSR}{**}          & 66.0 &59.2 &	62.4\\
PA-LSTM~\citep{Zhang2017PositionawareAA}     & 65.7 & \textbf{64.5} &	65.1\\
\midrule
GCN~\citep{Zhang2018GraphCO}                 & 69.8 &59.0 &	64.0\\
C-GCN~\citep{Zhang2018GraphCO}               & 69.9 &63.3 &	66.4\\
\midrule
AGGCN (ours)                                 & 69.9 & 60.9 &	65.1 \\
C-AGGCN (ours)                            & 73.1 & 64.2  & \textbf{69.0} \\
\bottomrule
\end{tabular}
\caption{Results on the TACRED dataset. Model with * indicates that the results are reported in~\citet{Zhang2017PositionawareAA}, while model with ** indicates the results are reported in~\citet{Zhang2018GraphCO}.  Our model with default random seed (0) achieves 69.0 F1 score. {The trained model is available for download, together with code and other details for reproducing the results. Following other practices in the community, we also conducted further experiments with 4 additional random seeds, and the averaged result over the 5 runs was: mean F1 68.2 with standard deviation 0.5.}}
\label{tab:tacred}
\end{table}

\begin{table}[!t]
	\centering
\setlength{\tabcolsep}{3pt}
\begin{tabular}{lcccc}
\toprule
\bf Model & & & &F1 \\
\midrule
SVM~\citep{Rink2010UTDCS}               & & & & 82.2\\
SDP-LSTM~\citep{Xu2015ClassifyingRV}    & & & & 83.7\\
SPTree~\citep{Miwa2016EndtoEndRE}       & & & & 84.4\\
PA-LSTM~\citep{Zhang2017PositionawareAA}& & & & 82.7\\
C-GCN~\citep{Zhang2018GraphCO}          & & & & 84.8\\
\midrule
C-AGGCN (ours)                           & & & &\textbf{85.7}    \\
\bottomrule
\end{tabular}
\caption{Results on the SemEval dataset. }
\label{tab:semeval}
\end{table}

As shown in Table \ref{tab:tacred}, the logistic regression classifier (LR) obtains the highest precision score. We hypothesize that the reason behind this is due to the data imbalance. This feature-based method tends to predict the relation to be the highly frequent labels (e.g., ``per:title''). Therefore, it has a high precision while has a relatively low recall. On the other hand, neural models achieve a better balance between precision and recall.

Since GCN and C-GCN already show their superiority over other dependency-based models and PA-LSTM, we mainly compare our AGGCN model with them.  We can observe that  AGGCN outperforms GCN by 1.1 F1 points. We speculate that the limited improvement is due to  the lack of contextual information about word order or disambiguation.  Similar to C-GCN~\citep{Zhang2018GraphCO}, we extend our AGGCN model with  a bi-directional LSTM network to capture the contextual representations which are subsequently fed into AGGCN layers. We term the modified model as C-AGGCN.  Our C-AGGCN model achieves a F1 score of 69.0, which outperforms the state-of-art C-GCN model.  We also notice that AGGCN and C-AGGCN achieve better precision and recall scores than GCN and C-GCN, respectively. The performance gap between GCNs with pruned trees and AGGCNs with full trees  empirically show that the AGGCN model is better at distinguishing relevant from irrelevant information for learning a better graph representation.





We also evaluate our model on the SemEval dataset under the same settings as~\citep{Zhang2018GraphCO}. Results are shown in Table~\ref{tab:semeval}. This dataset is much smaller than TACRED (only 1/10 of TACRED in terms of the number of instances). Our C-AGGCN model (85.7) consistently outperforms the  C-GCN model (84.8), showing the good  generalizability.

\subsection{Analysis and Discussion}

\begin{table}[!t]
	\centering
\setlength{\tabcolsep}{3pt}
\begin{tabular}{lcccc}
\toprule
\bf Model & & & &F1 \\
\midrule
C-AGGCN              & & & & 69.0\\
{\color{white}0}   -- Attention-guided layer (AG) &  & & & 67.1\\
{\color{white}0}   -- Dense connected layer (DC)  &  & & & 67.3 \\
{\color{white}0}   -- AG, DC   &  & & &66.7\\
{\color{white}0}   -- Feed-Forward layer (FF)    &  & & & 67.8\\
\bottomrule
\end{tabular}
\caption{An ablation study for C-AGGCN model. }
\label{tab:ablation}
\end{table}

\begin{table}[!t]
	\centering
\setlength{\tabcolsep}{3pt}
\begin{tabular}{lcccc}
\toprule
\bf Model & & & &F1 \\
\midrule
C-AGGCN (Full tree)             & & & & 69.0\\
C-AGGCN ($K$=2)             & & & & 67.5\\
C-AGGCN ($K$=1)             & & & & 67.9\\
C-AGGCN ($K$=0)             & & & & 67.0\\
\bottomrule
\end{tabular}
\caption{Results of C-AGGCN with pruned trees.}
\label{tab:prune_tree}
\end{table}

\begin{figure*}
    \centering
    \includegraphics[scale=0.52]{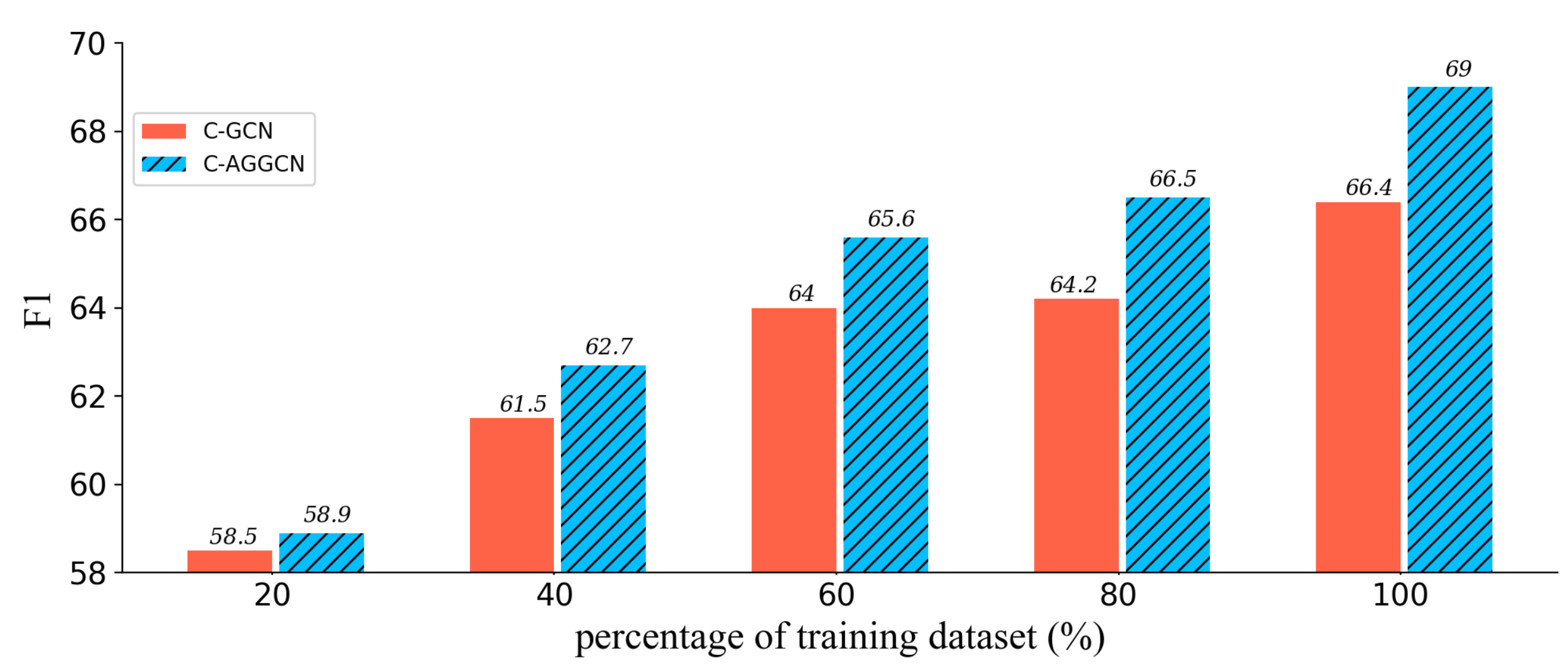}
    \caption{Comparison of C-AGGCN and C-GCN against different training data sizes. The results of C-GCN are reproduced from ~\cite{Zhang2018GraphCO}.}
    \label{fig:comp}
\end{figure*}

\begin{figure}[!t]
    \centering
    \includegraphics[scale=0.45]{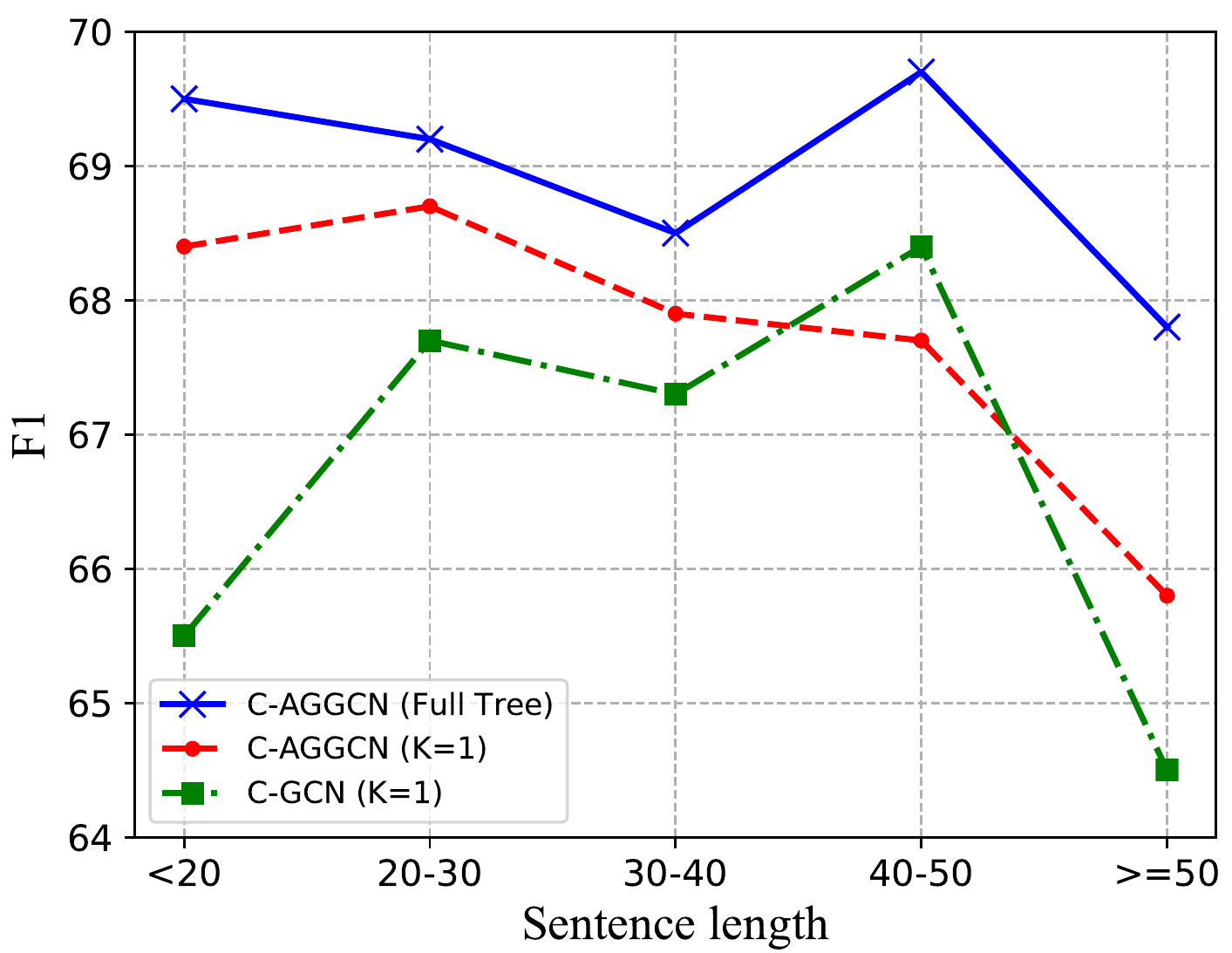}
    \caption{Comparison of C-AGGCN and C-GCN against different sentence lengths. The results of C-GCN are reproduced from ~\cite{Zhang2018GraphCO}.}
    \label{fig:tac_sl}
\end{figure}

\paragraph{Ablation Study.} We examine the contributions of two main components, namely, densely connected layers and attention guided layers, using the best-performing C-AGGCN model on the TACRED dataset.  Table~\ref{tab:ablation} shows the results. We can observe that adding either attention guided layers or densely connected layers improves the performance of the model. This suggests that both layers can assist GCNs to learn better information aggregations, producing better representations for graphs, where the attention-guided layer seems to be playing a more significant role.  We also notice that the feed-forward layer is  effective in our model. Without the feed-forward layer, the result drops to an F1 score of 67.8.

\paragraph{Performance with Pruned Trees.} Table~\ref{tab:prune_tree} shows the performance of the C-AGGCN model with pruned trees, where $K$ means that the pruned trees include tokens that are up to distance $K$ away from the dependency path in the  LCA subtree. We can observe that all the C-AGGCN models with varied values of $K$ are able to outperform the state-of-the-art C-GCN model~\citep{Zhang2018GraphCO} (reported in Table~\ref{tab:tacred}). Specifically,  with the same setting as $K$=1, C-AGGCN surpasses C-GCN by 1.5 points of F1 score. This demonstrates that, with the combination of densely connected layer and attention guided layer,  C-AGGCN  can learn better representations of graphs than C-GCN for downstream tasks. In addition, we notice that the performance of C-AGGCN with full trees outperforms all C-AGGCNs with pruned trees. These results further show  the superiority of ``soft pruning'' strategy   over hard pruning strategy in utilizing full tree information.

\paragraph{Performance against Sentence Length.} Figure~\ref{fig:tac_sl} shows the F1 scores of three models  under different sentence lengths.  We partition the sentence length into five classes ($<$ 20, $[$20, 30$)$, $[$30, 40$)$, $[$40, 50$)$,  $\geq$50). In general, C-AGGCN  with full trees outperforms C-AGGCN with pruned trees and C-GCN against various sentence lengths.  We also notice that C-AGGCN with pruned trees performs better than C-GCN in most cases. Moreover, the improvement achieved by C-AGGCN with pruned trees decays when the sentence length increases. Such a performance degradation can be avoided by using full trees, which provide more information of the underlying graph structures. Intuitively, with the increase of the sentence length, the dependency graph becomes larger as more nodes are included. This suggests that C-AGGCN can benefit more  from larger graphs (full tree).

\paragraph{Performance against Training Data Size.}
Figure~\ref{fig:comp} shows the performance of C-AGGCN and C-GCN  against different training settings. We consider five training settings (20$\%$, 40$\%$, 60$\%$, 80$\%$, 100$\%$ of the training data). C-AGGCN consistently outperforms C-GCN under the same amount of training data. When the size of training data increases, we can observe that the performance gap becomes more obvious. Particularly, using 80$\%$ of the training data, the C-AGGCN model is able to achieve a F1 score of 66.5, higher than C-GCN trained on the whole dataset. These results demonstrate that our model is more effective in terms of using training resources.

\section{Related Work}
\label{sec:4}

Our work builds on a rich line of recent efforts on relation extraction models and graph convolutional networks.

\paragraph{Relation Extraction.} 
Early research efforts are based on statistical methods. Tree-based kernels \citep{Zelenko2002KernelMF} and dependency path-based kernels \citep{Bunescu2005ASP} are explored to extract the relation. \citet{McDonald2005SimpleAF} construct maximal cliques of entities to predict relations. \citet{Mintz2009DistantSF} include syntactic features to a statistical classifier. Recently, sequence-based models leverages different neural networks to extract relations, including convolutional neural networks \citep{Zeng2014RelationCV, Nguyen2015RelationEP, Wang2016RelationCV}, recurrent neural networks \citep{Zhou2016AttentionBasedBL, Zhang2017PositionawareAA}， the combination of both \citep{Vu2016CombiningRA} and transformer \citep{Verga2018SimultaneouslyST}. 

Dependency-based approaches also try to incorporate structural information into the neural models. \citet{Peng2017CrossSentenceNR} first split the dependency graph into two DAGs, then extend the tree LSTM model \citep{Tai2015ImprovedSR} over these two graphs for $n$-ary relation extraction. Closest to our work, \citet{Song2018NaryRE} use graph recurrent networks \citep{Song2018AGM} to directly encode the whole dependency graph without breaking it. The contrast between our model and theirs is reminiscent of the contrast between CNN and RNN. Various pruning strategies have also been proposed to distill the dependency information in order to further improve the performance. \citet{Xu2015SemanticRC,Xu2015ClassifyingRV} adapt neural models to encode the shortest dependency path. \citet{Miwa2016EndtoEndRE} apply LSTM model over the LCA subtree of two entities. \citet{Liu2015ADN} combine the shortest dependency path and the dependency subtree. \citet{Zhang2018GraphCO} adopt a path-centric pruning strategy. Unlike these strategies that remove edges in preprocessing, our model learns to assign each edge a different weight in an end-to-end fashion.

\paragraph{Graph Convolutional Networks.} Early efforts that attempt to extend neural networks to deal with arbitrary structured graphs are introduced by \citet{Gori2005ANM, Bruna2014SpectralNA}. Subsequent efforts improve its computational efficiency with local spectral convolution techniques \citep{Henaff2015DeepCN, Defferrard2016ConvolutionalNN}. Our approach is closely related to the GCNs \citep{Kipf2016SemiSupervisedCW}, which restrict the filters to operate on a first-order neighborhood around each node. 


More recently, \citet{Velickovic2017GraphAN} proposed graph attention networks (GATs) to summarize neighborhood states by using masked self-attentional layers \citep{Vaswani2017AttentionIA}. Compared to our work, their motivations and network structures are different. In particular, each node only attends to its neighbors in GATs whereas AGGCNs measure the relatedness among all nodes. The network topology in GATs remains the same, while fully connected graphs will be built in AGGCNs to capture long-range semantic interactions.


\section{Conclusion}

We introduce the novel Attention Guided Graph Convolutional Networks (AGGCNs). Experimental results show that AGGCNs achieve state-of-the-art results on various relation extraction tasks. Unlike previous approaches, AGGCNs operate directly on the full tree and learn to distill the useful information from it in an end-to-end fashion. There are multiple venues for future work. One natural question we would like to ask is how to make use of the proposed framework to perform improved graph representation learning for graph related tasks \citep{Bastings2017GraphCE}.

\section*{Acknowledgements}

We would like to thank the anonymous reviewers for their valuable and constructive comments on this work. We would
also like to thank Zhiyang Teng, Linfeng Song, Yuhao Zhang and Chenxi Liu for their helpful suggestions. 
This work is supported by Singapore Ministry of Education Academic Research Fund (AcRF) Tier 2 Project MOE2017-T2-1-156. This work is partially supported by SUTD project PIE-SGP-AI-2018-01.

\bibliography{acl2019}
\bibliographystyle{acl_natbib}

\appendix
\clearpage

\section*{Supplemental Material}

\section{Case Study}
\label{sec:6}

Figure \ref{fig:Figure6} shows an instance in cross-sentence $n$-ary relation extraction task, which suggests that tumors with \textit{L858E} mutation in \textit{EGFR} gene partially responds to the drug \textit{gefitinib}. The relation between these three entities is sensitivity. We apply path-centric pruning on these dependency trees \citep{Zhang2018GraphCO}. The top one in Figure \ref{fig:Figure6} shows the pruned tree when $K$= 0 and the bottom one shows the pruned tree when $K$= 1. The state-of-the-art model, GCN over pruned tree ($K$= 0 and $K$= 1) \citep{Zhang2018GraphCO} predict the relation to be response rather than sensitivity. We hypothesize the reason is that the pruned tree miss the crucial information (i.e., partial response). The GCN model is not able to capture the interactions between removed tokens between entities, since these tokens are not in the resulting structure. 

\begin{figure}
    \centering
    \includegraphics[scale=0.5]{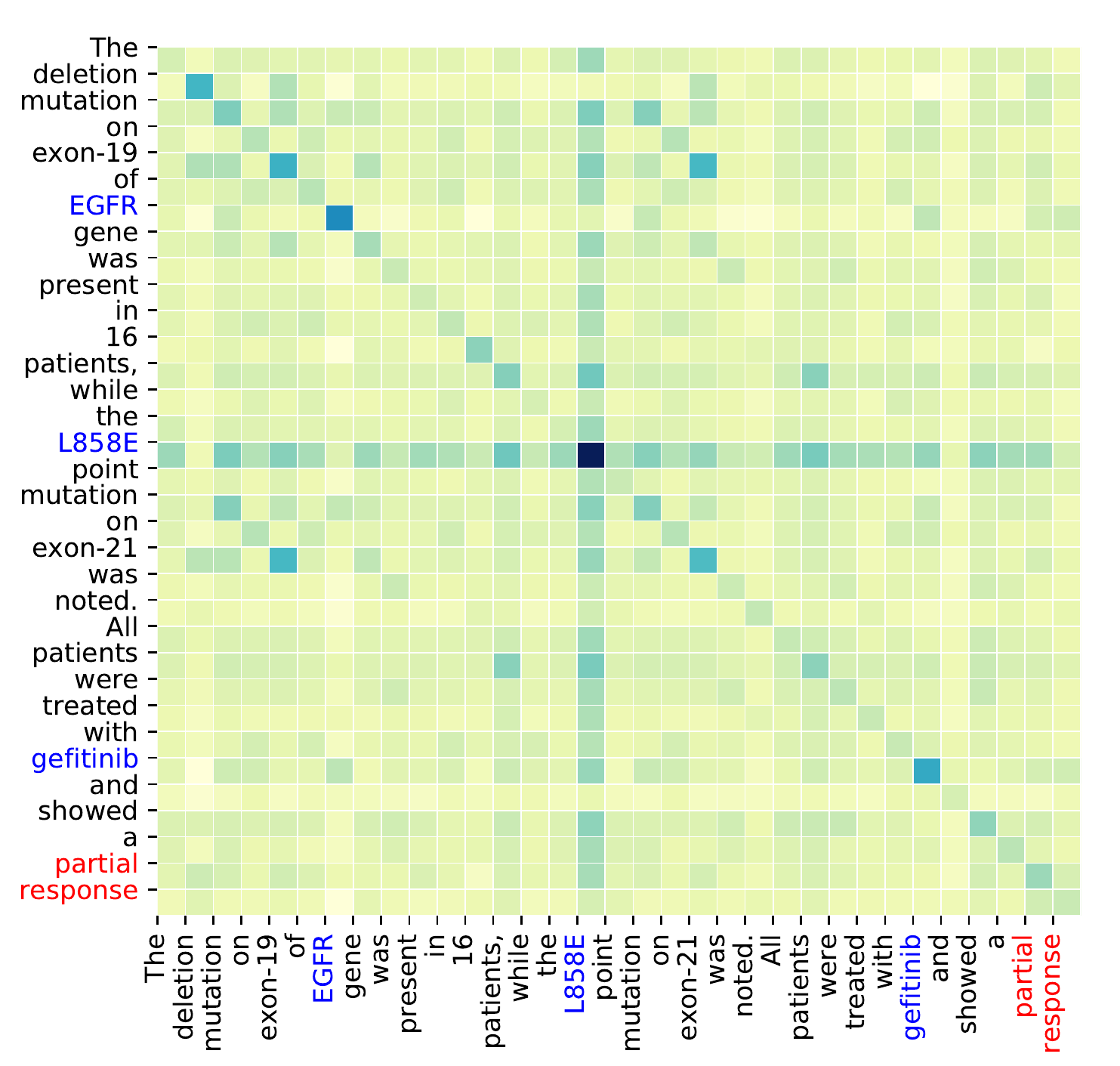}
    \includegraphics[scale=0.52]{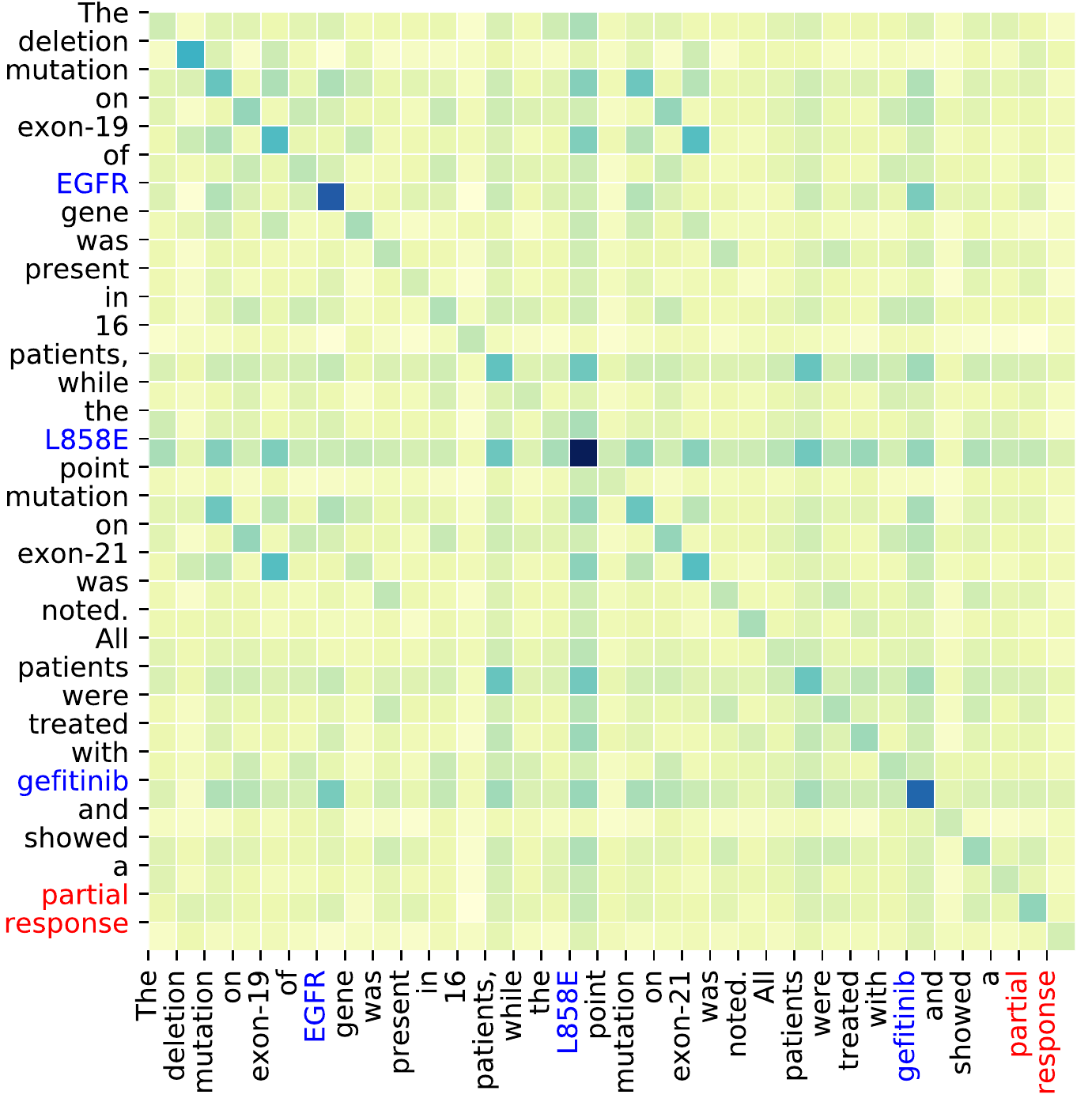}
    \caption{Visualizations of attention scores in the attention guided layer for the example in Figure \ref{fig:Figure6}. Darker color indicates higher score. Entities \textit{L858E}, \textit{EGFR} and \textit{gefitinib} are highlighted in blue. Tokens \textit{partial response} are highlighted in red. The top one shows the result of the first head and the bottom one shows the result of the second head.}
    \label{fig:Figure5}
\end{figure}

\begin{figure*}
    \centering
    \includegraphics[scale=0.5]{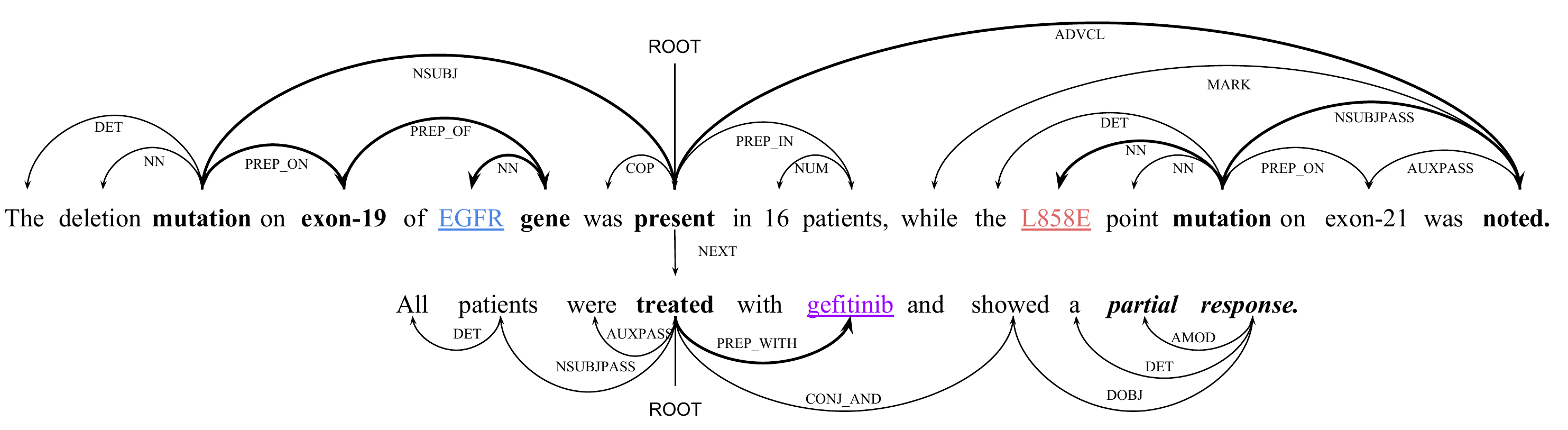}
    \includegraphics[scale=0.5]{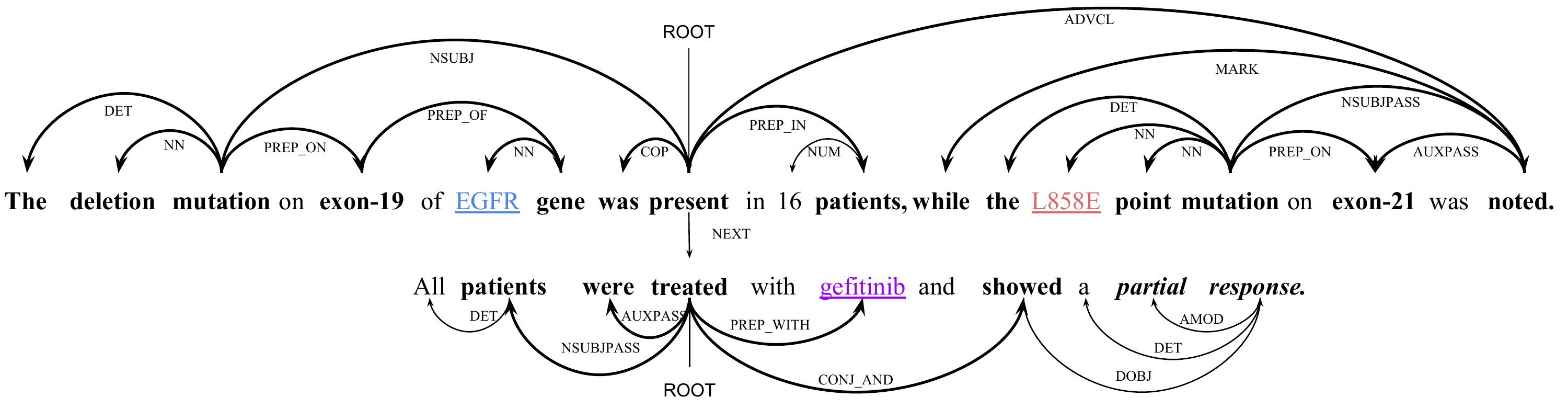}
    \vspace{-2mm}
    \caption{Example dependency trees for two sentences expressing a relation (sensitivity) among three entities \textit{L858E}, \textit{EGFR} and \textit{gefitinib}. The top one shows the pruned tree when $K$=0 (highlighted in bold). The bottom one shows the pruned tree when $K$=1 (highlighted in bold). Tokens \textit{partial response} are off these two paths.}
    \label{fig:Figure6}
\end{figure*}

Our AGGCN model predict the correct relation for this instance by including the attention guided layer, which is able to distill relevant information from the full tree in an end-to-end fashion. As shown in Figure \ref{fig:Figure5}, we visualize the attention scores of two heads in the attention guided layer. We can observe that relevant tokens including entity tokens and tokens that help to predict the correct relation (i.e., \textit{showed a partial  response}) can be attended by other tokens, especially the entity tokens. On the other hand, it is worthwhile to filter out noises from the dependency tree, especially when it is large. In the pruned tree shown in Figure \ref{fig:Figure6}, most stop words are removed directly in order to eliminate irrelevant information. In the AGGCN model, stop words including \textit{on, of, in, with, an, were} also have much lower scores. We believe the AGGCN model is able to maintain a balance between including and excluding information in the full tree for learning a better graph representation.

\section{Additional Analysis}
\label{sec:7}

\begin{figure}
    \centering
    \includegraphics[scale=0.4]{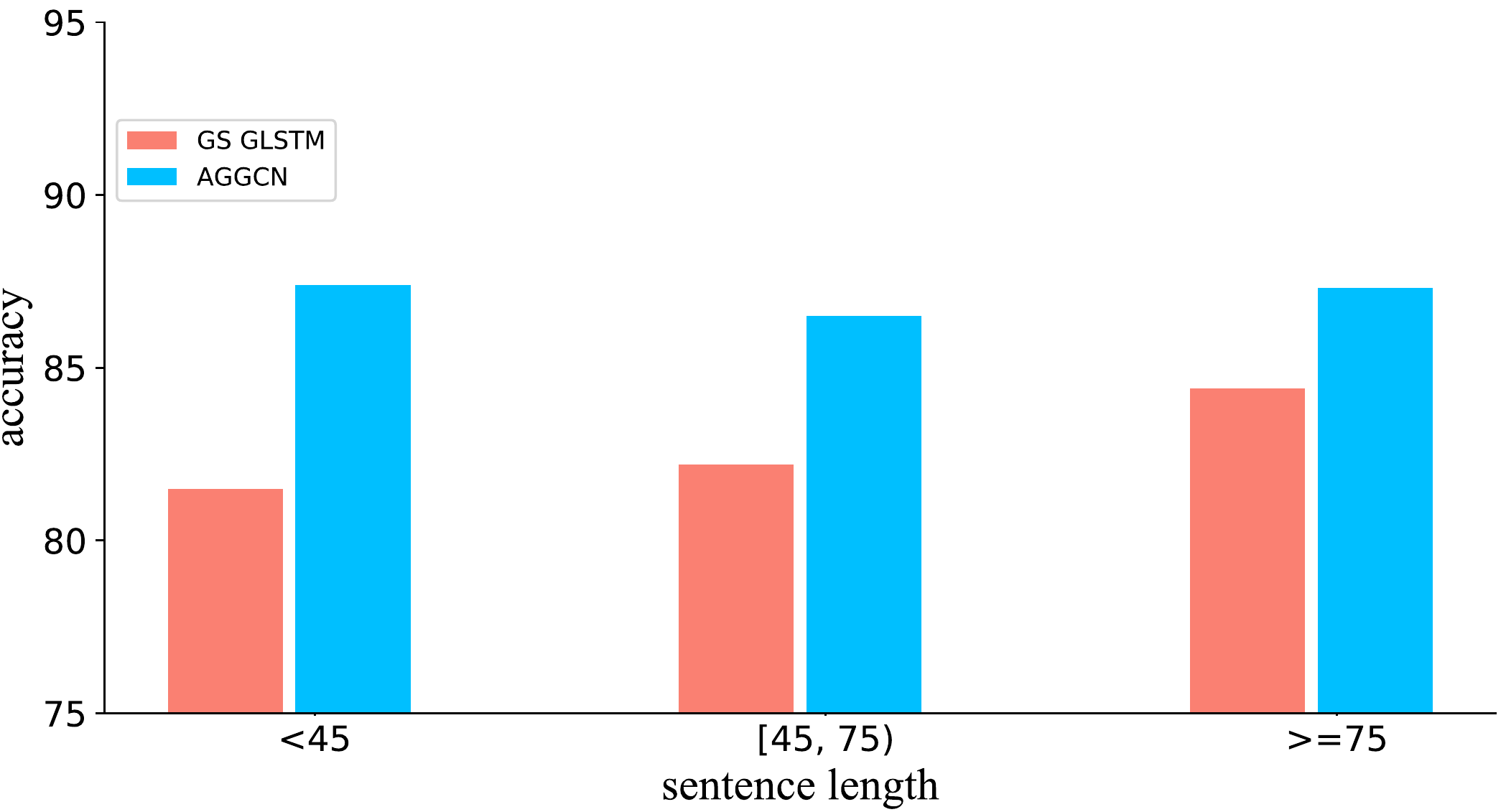}
    \includegraphics[scale=0.4]{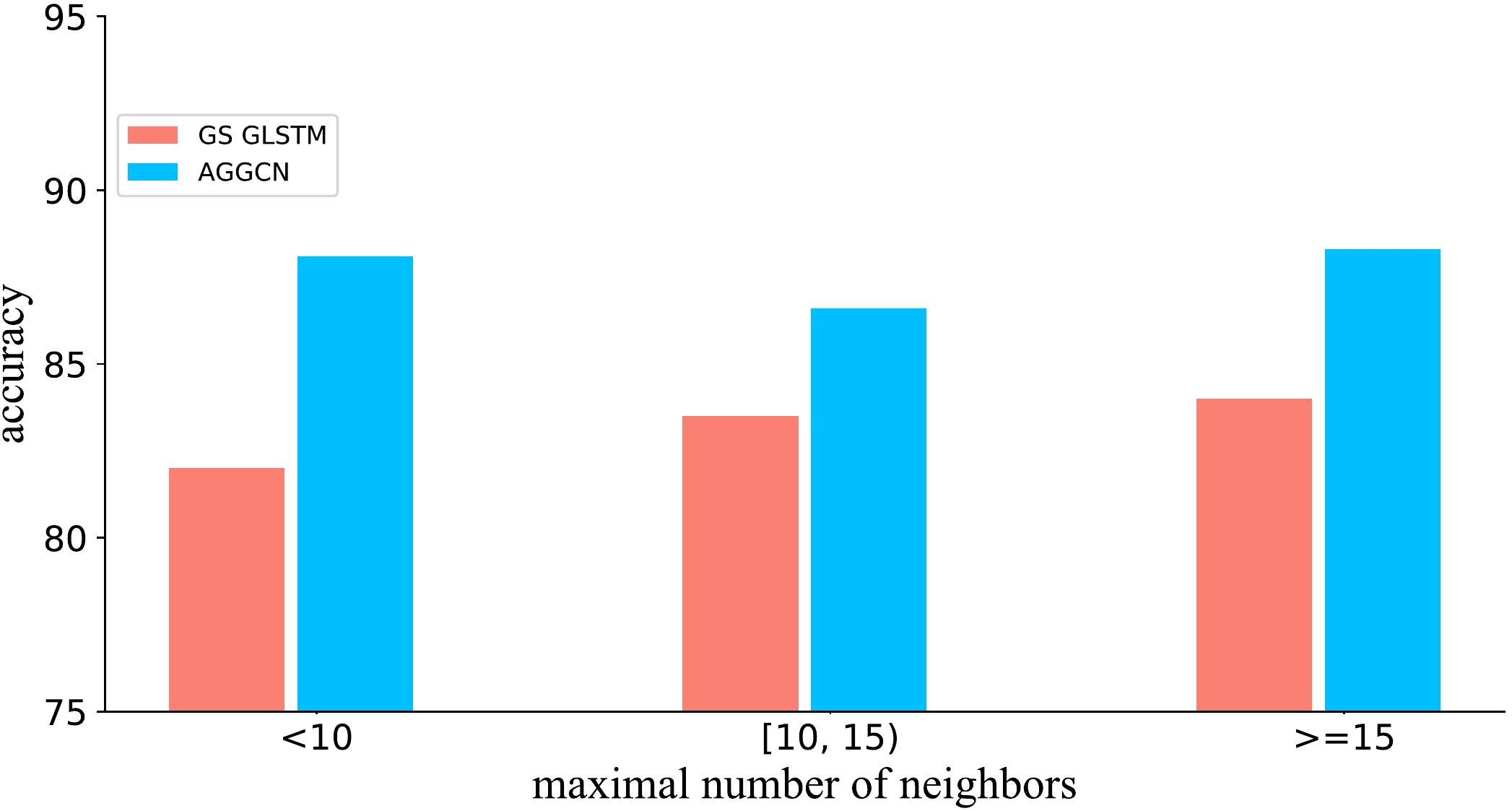}
    \caption{Test set performances of the AGGCN model and Graph State LSTM model (GS GLSTM). The top one shows performances on different sentence lengths. The bottom one shows performances on different maximal number of neighbors.}
    \label{fig:Figure7}
\end{figure}
\noindent
\citet{Song2018NaryRE} compare their Graph State LSTM model with the bidirectional DAG LSTM model against different sentence lengths and different maximal number of neighbors in order to better evaluate their model. Following the same setting, we compare our the test accuracies of the AGGCN model and Graph State LSTM  (GS GLSTM) under these two settings on cross-sentence $n$-ary relation extraction task as shown in Figure \ref{fig:Figure7}.

\textbf{Accuracy against sentence length}
We can observe that AGGCN consistently outperforms GS GLSTM. Specifically, the performance of GS GLSTM drops when the sentence is short, while the performance of AGGCN is stable. This shows that the superiority of AGGCN over GS GLSTM in utilizing context information.

\textbf{Accuracy against maximal number of neighbors}
Intuitively, it is easier to model graphs containing nodes with more neighbors, because these nodes can serve as a ``supernode'' that allow more efficient information exchange \citep{Song2018NaryRE}. The AGGCN model performs well under the inputs having lower maximal number of neighbors. We hypothesize that is because the attention guided layer convert the original graph into a fully connected graph, which encourages the information propagation.   

\end{document}